\documentclass[10pt, a4paper, conference, compsocconf]{IEEEtran}
\ifCLASSINFOpdf
\else
\fi

\usepackage{graphicx}
\usepackage{booktabs}

\begin{document}
%
\title{Hamtajoo: A Persian Plagiarism Checker for Academic Manuscripts}

\author{\IEEEauthorblockN{Vahid Zarrabi,
Salar Mohtaj, Habibollah Asghari}
\IEEEauthorblockA{ICT Research Institute \\ Academic Center for Education, Culture and Reseach (ACECR), Tehran, Iran \\
\{vahid.zarrabi $\mid$ salar.mohtaj $\mid$ habib.asghari\} @ictrc.ac.ir}}

\maketitle

\begin{abstract}
In recent years, due to the high availability of electronic documents through the Web, the plagiarism has become a serious challenge, especially among scholars. Various plagiarism detection systems have been developed to prevent text re-use and to confront plagiarism. Although it is almost easy to detect duplicate text in academic manuscripts, finding patterns of text re-use that has been semantically changed is of great importance. Another important issue is to deal with less resourced languages, which there are low volume of text for training purposes and also low performance in tools for NLP applications. In this paper, we introduce Hamtajoo, a Persian plagiarism detection system for academic manuscripts.
Moreover, we describe the overall structure of the system along with the algorithms used in each stage. 
In order to evaluate the performance of the proposed system, we used a plagiarism detection corpus comply with the PAN standards.
\end{abstract}

\IEEEpeerreviewmaketitle

\section{Introduction}
Plagiarism refers to the unacknowledged use of others' text or ideas. Stopping plagiarism is an important topic in the field of academic publication. Plagiarism detection systems compare a submitted suspicious document against a collection of source documents to find the cases of text re-use. It has been proved that plagiarism detection systems are effective tools for discouraging researchers to commit plagiarism \cite{BrownPreventing}. \par
There are two main categories for plagiarism detection (PD) systems. External plagiarism detection systems search for pattern of text re-use between suspicious document and a collection of source documents. On the other hand, intrinsic plagiarism detection systems exploit stylometry algorithms to find the changes in writing style of the author.  \par
There are different plagiarism detection tools that have been investigated in multiple surveys \cite{LancasterClassifications,LukashenkoComputer}. Crosscheck \cite{ZhangCrossCheck,Zhangsurvey} and Turnitin \cite{BataneTurning} are the most popular ones in academic community. Although there are many plagiarism detection systems in English, plagiarism detection is still a serious task in less resourced languages. \par
In this paper, we propose \textit{Hamtajoo}, a Persian plagiarism detection framework for investigating patterns of text re-use in Persian academic papers.
Persian belongs to Arabic script-based languages which share some common characteristics. This language family has some common properties such as absence of capitalization, right to left direction, encoding issues in computer environment and lack of clear word boundaries in multi-token words \cite{FarghalyComputer}. 
The system works on a document level at the first stage and then focuses of paragraph and sentence level in the second detailed comparison stage. We have prepared a reference collection of all of the Persian academic papers indexed by SID\footnote{  http://www.sid.ir/} (Scientific Information Database). \par
The rest of the paper is organized as follows. Section~\ref{relatedwork} describes the related work on recent plagiarism detection tools and methods. In section~\ref{proposedapproach}, we explain the proposed approach and also describe the algorithms that have been developed and implemented in the system. Section~\ref{systemarchitecture} comes with system design, architecture and functionalities. In section~\ref{experimentsandevaluation}, an evaluation framework for examining the system is described. Conclusion and the directions for the future work are presented in the final section. \par

\section{Related Work}
\label{relatedwork}
In this section we investigate some of the available plagiarism detection systems. \par
An Arabic plagiarism detection tool called Aplag has been introduced in \cite{MenaiAPlag}. For developing the plagiarism checker system, they have extracted the fingerprints on document, paragraph and sentence levels for saving the computation time. If the similarity between hashes of the two documents is above a specific threshold, then the process continues to paragraph level and so on. \par
In a work accomplished by Alzahrani et al., an intelligent plagiarism reasoned named as iPlag, has been designed \cite{AlzahraniiPlag}. Scientific publications in same fields usually share same general information and have some common knowledge. Besides, each publication should convey specific contributions. In this system, they have processed various parts of a manuscript and weighted them based on their importance in plagiarism detection. So, the PD system pays more attention to the parts of a manuscript that has more contributions and lower weights given to less important parts. \par
Meuschke et al. proposed Citeplag, a plagiarism detection system based on citation pattern matching \cite{MeuschkeCitePlag}. They search similar patterns of citations between source and suspicious documents to find cases of plagiarism. \par
In a work accomplished by Leong and Lau, they have developed a document plagiarism detection system named as Check \cite{SiCHECK}. They try to eliminate unnecessary comparisons between documents with different subjects and so reduced the computational costs. \par

A word-similarity sentence-based plagiarism detection tool on Web documents, named as SimPaD has been developed in \cite{PeraSimPaD}. They measure the similarity between sentences by computing word correlation factors and then generate a graphical view of sentences that are similar. \par
Collberg et al. developed a system for self-plagiarism detection named as SPLAT \cite{CollbergSPLAT}. The system crawls websites of top fifty computer science departments and downloads the research papers. Then a text comparison algorithm compares all of the papers for instances of text re-use. \par
It should be noted that most of the tools for detection cases of plagiarism are only pay attention to instances of verbatim copy plagiarism and cannot identify paraphrased passages of text which require semantic similarity detection methods. Our contribution in this paper is to use specific features in Persian to detect cases of paraphrased plagiarism.

\section{Proposed Approach}
\label{proposedapproach}
The proposed approach for developing \textit{Hamtajoo} system is thoroughly described in this section. \par
\begin{figure}[htbp]
    \centering
    \includegraphics[width=0.45\textwidth]{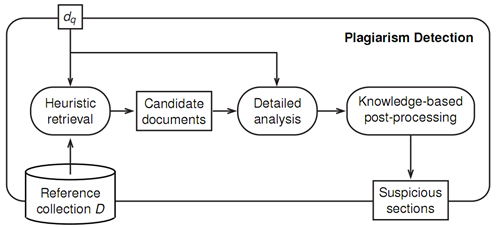}
    \caption{Generic three-step retrieval process for an external plagiarism detection system \cite{SteinStrategies}}
    \label{fig:1}
\end{figure}

Stein et al. proposed a generic three-step retrieval process for an external PD system \cite{SteinStrategies} that is depicted in Figure~\ref{fig:1}. Their proposed model consists of a heuristic retrieval step for extracting a set of source documents from the source collection, a detailed analysis step for comparing the suspicious document with candidate source documents for finding the similar passages, and a knowledge-based post-processing step to analyze identical passages and investigate whether they contain proper quotations. \par

A similar approach has been used to develop \textit{Hamtajoo} PD system. \textit{Hamtajoo} includes two main components, including the candidate retrieval and the text alignment modules. From the evaluation point of view, candidate retrieval is a recall-oriented task, while precision is mainly focused by the subsequent text alignment step \cite{PotthastOverview}. \par
In this section we describe main proposed methods for the candidate retrieval and the text alignment modules. The evaluation results on the proposed algorithms are presented in the next section. \par

\subsection{Candidate Retrieval Module}
Since the comparison of submitted suspicious document with the entire source documents in the system would be very time consuming, a candidate retrieval stage is considered to decrease the search space. The aim is to find most similar documents with the submitted suspicious document and also to reduce the number of documents for the subsequent text alignment stage. Our approach to retrieve candidate documents is divided into four main steps:

\begin{itemize}
    \item Chunking the suspicious document
    \item Noun phrase and keyword phrase extraction
    \item Query formulation
    \item Search control
\end{itemize}

Before these main steps, suspicious documents are passed through Parsiver pre-processing \cite{MohtajParsivar} block that includes stop words removal and unification of punctuation. Parsiver is an integrated package written in python which performs different kinds of pre-processing tasks in Persian. Each of the mentioned steps is described in more details as follows. \par
\textbf{Chunking the suspicious document:} In this step, the submitted suspicious document is segmented into some parts called chunks. These chunks will be used for query construction based on keyword phrase and noun phrase extraction. Therefore, their length should be long enough to extract meaningful queries. On the other hand, these chunks may contain unknown numbers of plagiarism cases from source documents. Based on experiments on \textit{Hamtajoo} to choose the system parameters, 500 words length has been chosen as the chunk length. In other words, the suspicious document would be divided into chunks of 500 words length and then each chunk is tokenized into individual sentences. \par
\textbf{Noun phrase and keyword phrase extraction:} This step has the main role in candidate retrieval task. In other words, extracting appropriate keywords could lead the candidate retrieval to perform more efficiently. There are many previous studies that tried to extract keywords by investigating content \cite{MatsuoKeyword,WittenKEA}. In order to construct queries from the suspicious document in our approach, both keyword phrases and noun phrases are extracted. \par
Before starting the extraction process, sentences with low information content are discarded. For this purpose, the input sentences have been ranked based on their length and the number of nouns, and then we discard the lower 20\% of the sentences. The resulting sentences are long enough and have rich content for keyword extraction. The TF-IDF (Term Frequency - Invert Document Frequency) weighting scheme has been used to extract important words from the sentences. The keywords are extracted from top 3 most important sentences that contain the highest TF-IDF words. The chosen keywords include nouns with high TF-IDF values, remaining nouns in the sentence, and also adjectives and verbs with high TF-IDF values. Moreover, the noun phrase extraction is accomplished by processing the remaining sentences. The formulation is deployed based on the formal Persian noun phrase structure. For each noun phrase, a score is calculated based on TF-IDF values. \par
\textbf{Query formulation:} For top ranked sentences selected from previous step, the extracted keywords are simply placed next to each other based on their order in sentence and are passed to the next step as a query. The Apache Lucene\footnote{https://lucene.apache.org/} is used to index the source documents. The constructed queries in this section are passed to Lucene application programming interface (API) to search them within the indexed documents. \par
\textbf{Search control:} In this step, some of the constructed queries are dropped based on the previously downloaded documents. The input query is compared against the downloaded documents that are gathered from previous rounds of source retrieval steps.  \par
The retrieved documents based on the constructed queries are passed to the text alignment sub-system for more detailed analysis and comparison. For each submitted documents, 25 source documents are chosen in the average for text alignment analysis in the next stage.

\subsection{Text Alignment Module}
The candidate documents from the previous stage are passed to a text alignment module for detail comparison of text passages between suspicious document and the source candidates. In this stage, the exact position of text re-use cases will be detected and will be reported to the end users. Different methods and algorithms have been tested to choose the best one from the accuracy and speed points of view. A standard PD corpus and also a plagiarism detection lab are designed to compare the performance of different methods. The detail description of the compiled corpus will be presented in the "Experiments and Evaluation" section. \par
\begin{figure}[b]
    \centering
    \includegraphics[width=0.44\textwidth]{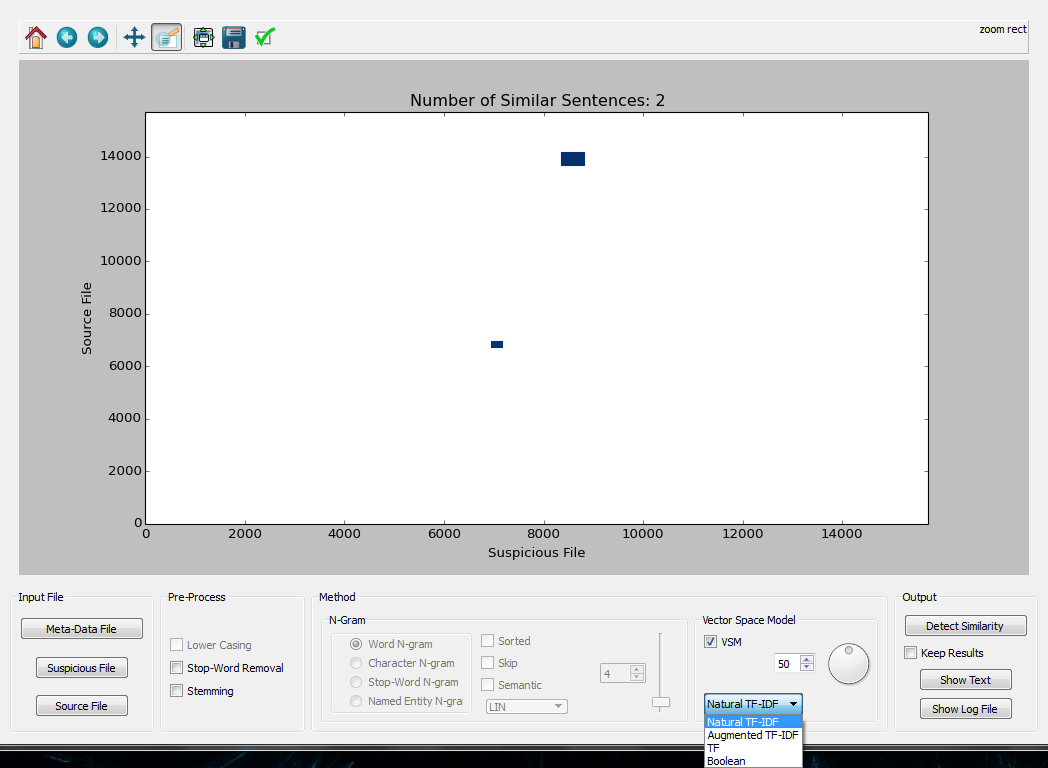}
    \caption{Snapshot of plagiarism detection lab}
    \label{fig:2}
\end{figure}

Figure~\ref{fig:2} shows a snapshot of our PD lab. As depicted in the figure, different methods including character n-gram, word n-gram, Vector Space Model (VSM) and Latent Semantic Analysis (LSA) are deployed in the tool. Moreover, different parameters such as the range of n for n-gram similarity, the similarity threshold for VSM and LSA and the parameters related to different parts of pre-processing are embedded in our PD lab to analyze their impact on plagiarism detection performance.

As depicted in the figure, the PD lab includes a dot plot graph that shows the cases of similarity between pairs of documents. Among different methods which have been developed in our PD lab, the VSM similarity is chosen based on accuracy and runtime criteria. \par
The proposed model for detecting exact position of text re-use cases includes the following steps; First, we split the pairs of suspicious and candidate source documents into sentences. In the second step, for each sentence in source and suspicious document, the relevant vectors have been created based on TF-IDF weighting schema. The IDF measure has been computed on a large collection of academic manuscript and the TF counts the number of target word in the document. Finally, in the third step, a pairwise cosine similarity has been computed between all of the sentences in two documents.  Pairs of the sentences with similarity higher than predefined threshold are considered as cases of text re-use.

\section{System Architecture}
\label{systemarchitecture}
\begin{figure}[t]
    \centering
    \includegraphics[width=0.25\textwidth]{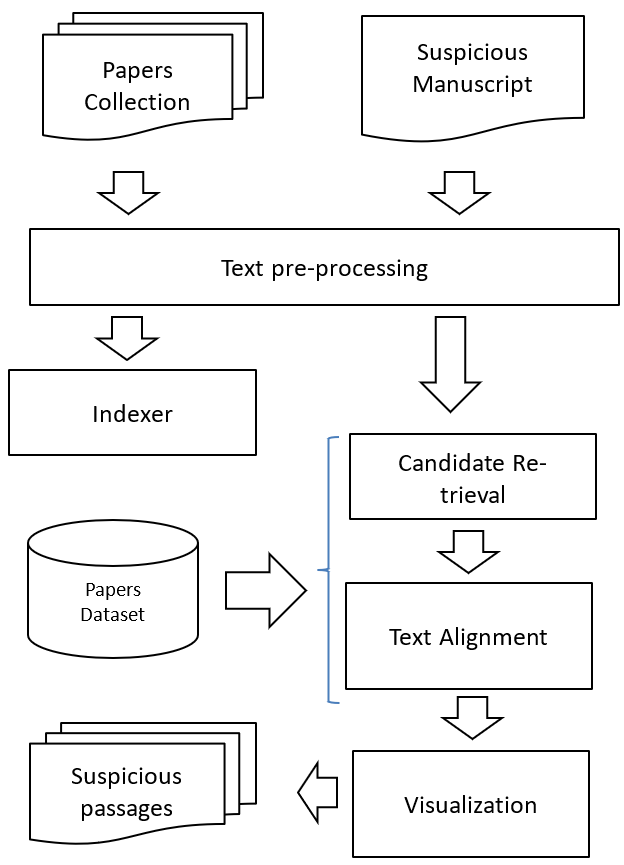}
    \caption{The block diagram of \textit{Hamtajoo} PD system}
    \label{fig:3}
\end{figure}

In this section, we describe the main technologies that are used to develop the PD system. \textit{Hamtajoo} is a web application composed of two main subsystems; the core (back-end) and front-end subsystems. The main stages of \textit{Hamtajoo} are depicted in Figure~\ref{fig:3}.

\subsection{Core (back-end) Sub-system}
The Django (1.8) web framework has been used to develop the core subsystem of \textit{Hamtajoo}. Django is a free and open-source web framework, written in Python, which follows the model-view-template (MVT) architectural pattern . The user authentication, raw text extraction from submitted documents and the candidate retrieval and text alignment tasks are done in the core subsystem. \par
To make a standard format from both submitted suspicious document and the source documents in the system, the Parsiver pre-processing toolkit \cite{MohtajParsivar} is used. Parsiver has been used to normalize all of the character encodings into a unified format and also to tokenize words in text documents. Moreover, we used Parsiver stemmer for different functions in text alignment module of the system. \par
The input documents to \textit{Hamtajoo} can be of various file formats including ".doc", ".docx" and ".txt" formats. To extract the raw text from .doc and .docx documents, the win32com.client and docx2txt modules are used, respectively. The docx2txt is a pure python-based utility to extract text from .docx files. The code is taken and adapted from python-docx\footnote{https://github.com/ankushshah89/python-docx2txt}. We have used these modules to extract the body of text from the submitted documents. \par
In order to construct the collection of source documents, all of the papers from SID scientific database were fed into the \textit{Hamtajoo} system. To do so, the Apache Lucene platform has been used to index the text documents. While the python is used to develop the core subsystem, the PyLucene API is used to develop the indexing system. PyLucene is a Python extension for accessing Java Lucene. Its goal is to allow using Lucene text indexing and searching capabilities from Python\footnote{https://lucene.apache.org/pylucene/index.html}. Moreover, MySQL database is employed to store the metadata information (e.g. author, title and publishing year of paper) that are extracted from indexed papers in the system.

\subsection{Front-end Sub-system}
The front-end subsystem contains user interface which makes it possible for end users to use \textit{Hamtajoo} for investigating plagiarism in their documents. Moreover, users can create and manage user accounts using the system interface. We exploited the Bootstrap web framework to develop the front-end subsystem. Bootstrap is a free and open-source front-end Web framework for designing websites and Web applications. It contains HTML- and CSS-based design templates for typography, forms, buttons, navigation and other interface components, as well as optional JavaScript extensions. \par
\begin{figure}[!b]
    \centering
    \includegraphics[width=0.46\textwidth]{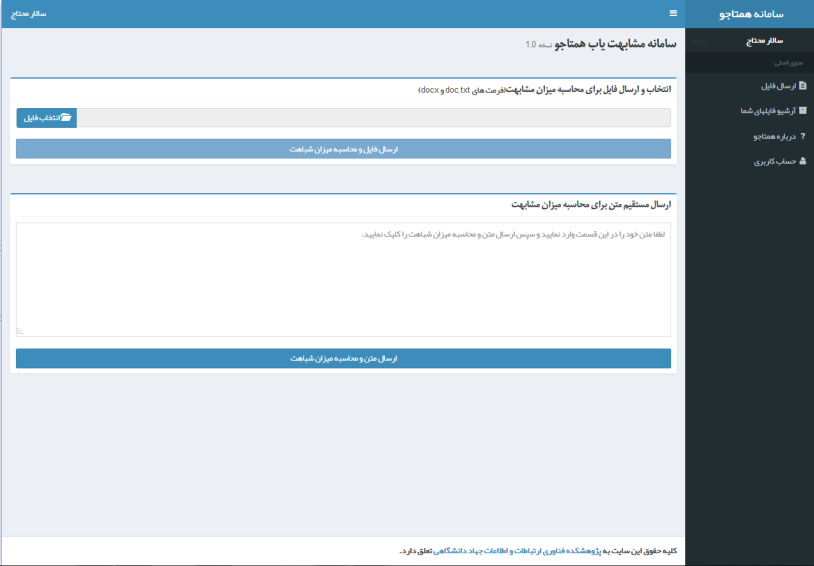}
    \caption{Manuscript submission page of \textit{Hamtajoo}}
    \label{fig:4}
\end{figure}

Figure~\ref{fig:4} shows the main parts of front-end section of \textit{Hamtajoo}. This figure shows the main page of the system, where users can submit their documents for plagiarism analysis. It is possible for users to submit the raw text directly into the system and to upload their documents in doc, docx and txt formats. \par

\begin{figure}[!t]
    \centering
    \includegraphics[width=0.46\textwidth]{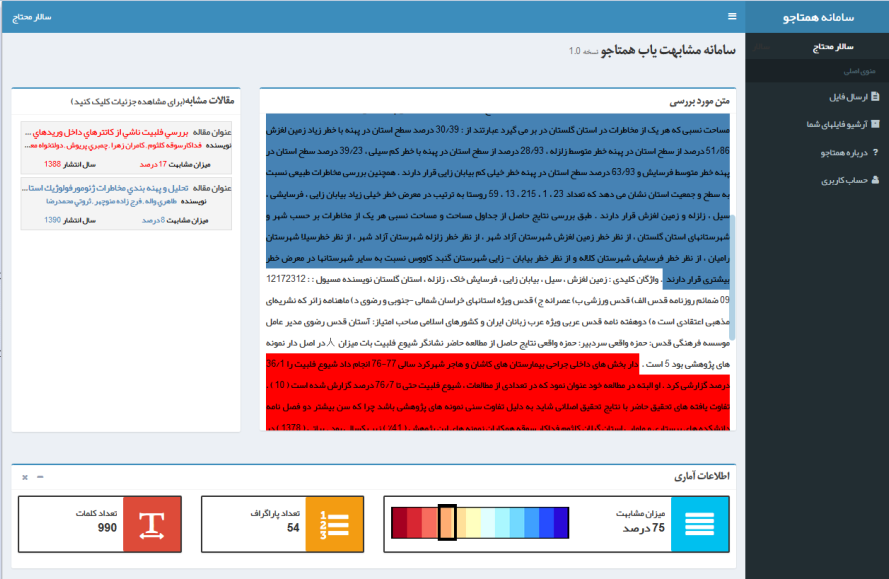}
    \caption{System output for detail comparison of source and suspicious documents}
    \label{fig:5}
\end{figure}

Fig~\ref{fig:5} shows the system output for a submitted suspicious document. The system highlights the section of the submitted text which contains cases of text re-use with different colors for different source papers. Moreover, some general statistical information related to submitted document (i.e. number of words, number of paragraphs and ratio of plagiarism in the document) are depicted in the bottom side of the page. List of the source papers which have cases of text similarity with the submitted document can be shown in system output (two papers in this example). Users can examine the detail text similarity between the submitted text and source papers by clicking on listed papers.

\begin{table*}
    \caption{Overall detection performance of \textit{Hamtajoo} vs. the algorithms presented in Persian PlagDet 2016 \cite{AsghariAlgorithms}}
    \label{tab:1}
    \centering
    \begin{tabular}{l|c|c|c|c|c}
        \toprule
        Team & Recall & Precision & Granularity & F-Measure & Plagdet \\
        \midrule
            \textbf{Hamtajoo} & \textbf{0.9221} & \textbf{0.9345} & \textbf{1} & \textbf{0.9282} & \textbf{0.9282} \\
            Mashhadirajab & 0.9191 & 0.9268 & 1.0014 & 0.9230 & 0.9220 \\
            Gharavi & 0.8582 & 0.9592 & 1 & 0.9059 & 0.9059 \\
            Momtaz & 0.8504 & 0.8925 & 1 & 0.8710 & 0.8710 \\
            Minaei & 0.7960 & 0.9203 & 1.0396 & 0.8536 & 0.8301 \\
            Esteki & 0.7012 & 0.9333 & 1 & 0.8008 & 0.8008 \\
            Talebpour & 0.8361 & 0.9638 & 1.2275 & 0.8954 & 0.7749 \\
            Ehsan & 0.7049 & 0.7496 & 1 & 0.7266 & 0.7266 \\
            Gillam & 0.4140 & 0.7548 & 1.5280 & 0.5347 & 0.3996 \\
            Mansourizadeh & 0.8065 & 0.9000 & 3.5369 & 0.8507 & 0.3899 \\
        \bottomrule
    \end{tabular}
\end{table*}

\section{Experiments and Evaluation}
\label{experimentsandevaluation}
In order to evaluate \textit{Hamtajoo} plagiarism detection system, two different experiments have been accomplished to measure the performance of different subsystems. Our candidate retrieval module has been evaluated in PAN 2015 international competition on plagiarism detection \cite{HagenSource}. The results are presented in the following subsection. To evaluate the text alignment module, the standard PD dataset of Persian PlagDet shared tasks \cite{AsghariAlgorithms} on plagiarism detection has been used. The performance of text alignment module is compared with all of the proposed methods in Persian PlagDet competition.

\subsection{Candidate Retrieval Evaluation}
As mentioned in the previous section, in order to evaluate the proposed candidate retrieval approach, we participated in the source retrieval task of PAN 2015 international shared task on plagiarism detection. The candidate retrieval module of \textit{Hamtajoo} achieved the best results in "runtime" and "No detection" measures. Moreover, \textit{Hamtajoo} achieved the second rank in recall measure and also the number of queries among all of the participants. The results of source retrieval shared task are presented in detail in \cite{HagenSource}.

\subsection{Text Alignment Evaluation}
The Persian PlagDet shared task at PAN 2016 \cite{AsghariAlgorithms} has been organized  to promote the comparative assessment of NLP techniques for plagiarism detection with a special focus on plagiarism that appears in a Persian text corpus. Since the shard task was focused on Persian, we have evaluated the performance of \textit{Hamtajoo} using standard Persian PlagDet 2016 evaluation corpus. Table~\ref{tab:1} shows the performance results of \textit{Hamtajoo} in comparison to participants of Persian PlagDet 2016. As mentioned in the table, \textit{Hamtajoo} outperforms other systems in different evaluation measures.

\section{Conclusion and Future Works}
In this paper, we introduced \textit{Hamtajoo}, a Persian plagiarism detection system. It has been built around a semantic-based method considering specific features of Persian language. The system contains a resource collection comprised of about 480000 journal papers in Persian. Pre-processing of text documents was done in order to transform them into a unified representation, normalizing the text (e.g. unification of various character encodings) and sentence/word tokenization. \par
By exploiting a graph showing the distribution of plagiarized passages across the document, the expert could achieve a better view of re-used text. The experimental results show the effectiveness of \textit{Hamtajoo} and its competitiveness against the other plagiarism checker tools. For adapting the system to a commercial package, a web-based system is developed to present the system to the academic society. \par
As a plan for the future works, we aim to conduct more experiments on larger texts to detect the text re-use in long documents such as theses and dissertations with a reasonable computational complexity. Another additional work is to deal with bilingual algorithms for detecting plagiarized passages translated from English to Persian. Further improvements can also be done by integrating \textit{Hamtajoo} and \textit{Maglet} \cite{MohtajMaglet} that is a Persian journal recommender system. It can facilitate the manuscript submission process for academicians by checking plagiarism and also by finding appropriate journals for their manuscripts in an integrated system. 

\section*{Acknowledgment}
This work has been accomplished in ICT research institute, ACECR and funded by Vice Presidency for Science and Technology of Iran - Grant No. 1164331. We would like to thank all of the members of ITBM and AIS research groups of ICT research institute for their contribution in corpus construction. Especial credit goes to Javad Rafiei and Khadijeh Khoshnava for their help in development and testing the algorithms.



%

\end{document}